\documentclass{article}

\usepackage[final]{neurips_2025}

\usepackage[utf8]{inputenc} 
\usepackage[T1]{fontenc}    
\usepackage{hyperref}       
\hypersetup{
    colorlinks=true,
    urlcolor=magenta,  
}
\usepackage{url}            
\usepackage{booktabs}       
\usepackage{amsfonts}       
\usepackage{nicefrac}       
\usepackage{microtype}      
\usepackage{xcolor}         
\usepackage{graphicx}
\usepackage{caption}
\usepackage{subcaption}
\usepackage{amsmath}
\usepackage{pifont}
\usepackage{listings} 
\title{RigAnyFace: Scaling Neural Facial Mesh Auto-Rigging with Unlabeled Data}

\author{
Wenchao Ma$^{1}$\thanks{Equal Contribution.} \thanks{Work partially completed during an internship at Roblox.} 
\quad Dario Kneubuehler$^{2}$\footnotemark[1]
\quad Maurice Chu$^{2}$ 
\quad  Ian Sachs$^{2}$ \\ 
\quad \textbf{Haomiao Jiang}$^{2}$ 
\quad \textbf{Sharon X. Huang}$^{1}$ \\ \
$^{1}$Penn State University \quad $^{2}$Roblox
}
\begin{document}
\maketitle
\begin{center}
\vspace{-6mm}
    \captionsetup{type=figure}
    \includegraphics[width=0.8\linewidth]{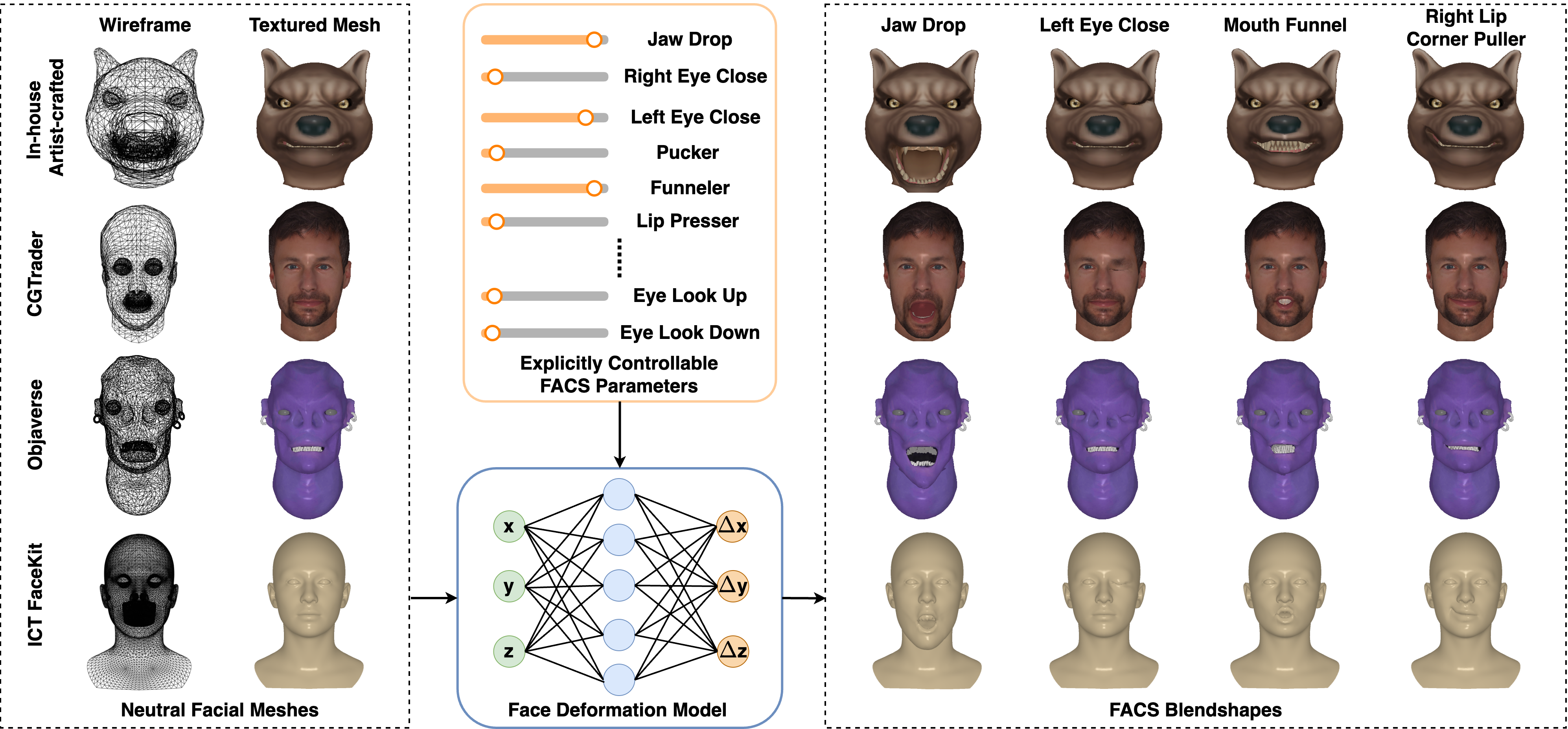}
    \captionof{figure}{We present RigAnyFace (RAF), an auto-rigging framework that supports facial meshes of diverse topologies with multiple disconnected components such as eyeballs. These meshes are drawn from diverse sources and cover both humanoid and non-humanoid heads. Given only a neutral facial mesh and explicitly controllable FACS parameters specifying activated action units, RAF accurately deforms the input mesh into corresponding FACS poses, creating an expressive blendshape rig.}
    \label{fig:teaser}
\end{center}
\begin{abstract}
In this paper, we present RigAnyFace (RAF), a scalable neural auto-rigging framework for facial meshes of diverse topologies, including those with multiple disconnected components. RAF deforms a static neutral facial mesh into industry-standard FACS poses to form an expressive blendshape rig. Deformations are predicted by a triangulation-agnostic surface learning network augmented with our tailored architecture design to condition on FACS parameters and efficiently process disconnected components. For training, we curated a dataset of facial meshes, with a subset meticulously rigged by professional artists to serve as accurate 3D ground truth for deformation supervision. Due to the high cost of manual rigging, this subset is limited in size, constraining the generalization ability of models trained exclusively on it. To address this, we design a 2D supervision strategy for unlabeled neutral meshes without rigs. This strategy increases data diversity and allows for scaled training, thereby enhancing the generalization ability of models trained on this augmented data. Extensive experiments demonstrate that RAF is able to rig meshes of diverse topologies on not only our artist-crafted assets but also in-the-wild samples, outperforming previous works in accuracy and generalizability. Moreover, our method advances beyond prior work by supporting multiple disconnected components, such as eyeballs, for more detailed expression animation.\textit{ Project page:} \href{https://wenchao-m.github.io/RigAnyFace.github.io}{https://wenchao-m.github.io/RigAnyFace.github.io}

\end{abstract}
\section{Introduction}
Facial rigging aims to make a static neutral facial mesh animatable by defining a set of controllable deformations, typically represented either as blendshape rigs driven by activated action units in FACS-based systems~\cite{paul1978facial, Lewis2014pratice, lewis2010direct, seol2011artist, cetinaslan2020stabliized, cetinasla2020sketching, NFR} or as skeletal rigs driven by joint positions~\cite{Vesdapunt2020JNR, kavan2024compress, orvalho2008transfering}. This is an essential step for creative AI, bringing digital avatars to life by enabling expressive and realistic facial movements across a wide range of applications. However, creating a rig for facial animation is laborious and expensive, often requiring skilled artists tens of hours to complete a single asset. In this paper, we propose a fully automated and generalizable facial rigging framework that alleviates the reliance on manual labor while achieving high-quality facial rigging.

Typically, facial auto-rigging methods transfer a complete set of blendshapes from a predefined template mesh to a neutral target facial mesh, often necessitating dense correspondences~\cite{NohN2001noh,sumner2004deformation, hao2010example, Prashanth2022Local} or a fixed mesh topology between the template and the target~\cite{jiaman2020dynamic,carrigan2020expression}.  Recent approaches~\cite{wang2023fully, chandran2022shape} utilize per-face VQ-VAEs~\cite{wang2023fully} to build transferable latent spaces between faces or triangulation-agnostic networks~\cite{chandran2022shape} to bypass these limitations. However, a template blendshape rig is still required, which can compromise accuracy when the template and target shapes differ substantially. NFR~\cite{NFR} is currently the only approach capable of directly rigging facial meshes from explicitly controllable Facial Action Coding System (FACS)~\cite{paul1978facial} parameters without relying on a template, although it has so far been demonstrated primarily on humanoid heads. Furthermore, existing approaches, including NFR, have yet to accommodate meshes with multiple disconnected components, such as eyeballs or teeth, limiting their ability to animate highly expressive avatars; for example, an ``eye lookdown'' pose is difficult to reproduce if the mesh lacks eyeballs.


To address the above challenges, we aim to build a facial auto-rigging framework with the following advantages: \textbf{(i)} it eliminates the reliance on predefined template blendshapes, removing the constraint that target facial meshes must rigorously resemble a predefined template; \textbf{(ii)} it is capable of animating in-the-wild facial meshes with varying topologies and shapes, including humanoid and non-humanoid samples as shown in Fig~\ref{fig:teaser}; and \textbf{(iii)} it supports facial meshes with multiple disconnected components to enable realistic and expressive 3D face animations. 

We present RigAnyFace (RAF), a scalable and generalizable framework for facial auto-rigging. RAF employs a facial mesh deformation network built on DiffusionNet~\cite{DiffusionNet}, a triangulation-agnostic backbone for meshes of different topologies. Guided by explicitly controllable FACS parameters, this network deforms a neutral facial mesh into a predefined set of FACS poses to form a blendshape rig. Compared to the original DiffusionNet, we introduce two key modifications: (i) a conditional diffusion block that extends the original diffusion block to incorporate FACS parameters as additional conditional inputs, and (ii) a global encoder designed to capture holistic mesh characteristics, enabling effective handling of multiple disconnected components. For network training, we curated a comprehensive dataset of facial meshes encompassing a wide variety of shapes with detailed disconnected components such as eyeballs and teeth. A subset of these meshes was meticulously rigged by professional artists to provide accurate ground-truth for 3D deformations. 

Relying solely on rigged heads for training limits the model’s generalizability in practice, given the scarcity of rigged samples due to the high cost of manual rigging. This motivates us to employ 2D supervision, which offers better accessibility and broader scalability compared to 3D supervision. We developed a 2D supervision strategy for 3D facial mesh deformation models, integrating appearance guidance from RGB images for prominent facial expressions and motion guidance from optical flow-like 2D displacement field for subtle micro-expressions. Supported by a generative 2D face animation model that synthesizes posed images from the renderings of a neutral mesh, along with an optical flow estimator that predicts the 2D displacement between neutral and posed images as 2D supervisions, we expand the training dataset using unlabeled neutral meshes without rigs. This enables the network to effectively distill rigging knowledge across diverse facial shapes, resulting in more accurate and generalizable 3D facial animations even with limited labeled training data.

Experiments show that our method outperforms prior work across assets from diverse sources, including our artist-crafted meshes and in-the-wild models from ICT FaceKit~\cite{ictface}, Objaverse~\cite{Objectverse}, and CGTrader~\cite{cgtrader2025}. In addition, we demonstrate several downstream applications of our auto-rigging system in user-controlled animation, retargeting human expressions from videos, and rigging generated facial meshes from a text-to-3D model.

\section{Related Works}
\textbf{Auto-rigging.} Auto-rigging facilitates efficient and realistic animation of 3D models by automatically generating hierarchical control systems. For full-body character auto-rigging~\cite{ilya2007automatic,alec2011bounded,ladislav2012elasticity, peizhuo2021learning,lijuan2019neuroskinning,jing2023tarig,albert2022skinningnet,zhan2020rignet,zhiyang2024make,song2025magicarticulate,liu2025riganything,zhang2025unirig, deng2025anymate}, most approaches follow a two-step pipeline: skeleton construction and skinning to generate the Linear Blend Skinning (LBS) rig. In contrast, facial character rigs are often anatomically-inspired, typically based on the Facial Action Coding System (FACS) ~\cite{paul1978facial}, a standardized framework that describes facial movements as combinations of muscle activations and is primarily implemented using blendshapes~\cite{Lewis2014pratice, lewis2010direct,seol2011artist,cetinaslan2020stabliized}.

Previous facial auto-rigging works~\cite{NohN2001noh,sumner2004deformation, hao2010example, Prashanth2022Local, jiaman2020dynamic,carrigan2020expression,wang2023fully,chandran2022shape} are mostly based on a complete set of blendshapes from a predefined template mesh, transferring the template blendshapes to the target mesh. For example, Li et al.\cite{jiaman2020dynamic} proposed a CNN-based approach that predicts offsets between template and target blendshapes represented by 2D geometry images. Chandran et al.~\cite{chandran2022shape} use a transformer with positional encodings to map meshes into a canonical space from user-marked correspondences, enabling deformation transfer from template to target across different topologies. Several notable works~\cite{victoria2019aDecoupled,chandran2020semantic,zhang2023dreamface,cao2014,tianye2017learning,anurag2018generating,jiang2019disentangled} can directly generate animatable 3D faces based on 3D Morphable Models~\cite{blanz1999amorphable}. NFR \cite{NFR} is able to deform a neutral facial mesh into target expressions by decoding FACS-aligned latent codes from a mesh auto-encoder, eliminating any template requirement. Its triangulation-agnostic backbone, trained on several face-animation datasets, generalizes to in-the-wild meshes with diverse topologies. Compared with NFR and other previous works, our method enables 2D supervision for scaled training and further improves accuracy and generalizability across a wider variety of facial meshes while natively supporting multiple disconnected components to allow finer-grained and more realistic expression animation.

\textbf{Facial Animation Transfer from 2D.} Facial animation transfer aims to retarget facial expressions from one character to another. Recent methods (e.g., ~\cite{free_avatar,aneja2018learning,Byungkuk2022Animatomy,larey2023facial,kim2021deep,moser2021semi}) show impressive results in transferring expressions to 3D avatars from 2D images or videos. However, these methods focus on transferring expressions to avatars that already have a rig and are hence not directly comparable to our work, which focuses on automatically generating rigs for facial meshes.

Significant progress has also been made in transferring facial animation for both single‑view and multi‑view images and videos~\cite{jianzhu2024liveportrait,shurong2024megactor,tingchun2021one-shot,you2024xportrait, nikita2024emoportrait, yue2024follow,tran2024voodoo,li2023generalizable, deng2024portrait4d, deng2024portrait4dv2,tran2024voodooxp,kartik2024gaussianheads}. Given a reference identity image, these methods can generate and manipulate facial expressions for the given identity using various control inputs, such as posed images of other identities or landmarks. Recent advancements in generative models~\cite{Diffusion, ControlNet} and the availability of large-scale face video datasets~\cite{arsha2017voxceleb,hao2022celebv} have enabled those methods to achieve remarkable success in 2D facial expression animation. For instance, MegActor\cite{shurong2024megactor} utilizes a diffusion-based generative framework, incorporating a motion disentanglement module to separate identity and expression features, and a motion retargeting model to map expressions onto target portraits. In this work, we utilize 2D face animation models to generate 2D supervision for unrigged heads. Our proposed framework is agnostic to the choice of 2D face animation model, provided they deliver satisfactory animation results. In practice, we base our 2D supervision generation on MegActor~\cite{shurong2024megactor}, which is open-source and efficient to fine-tune.

\section{Preliminary}
\subsection{DiffusionNet}

DiffusionNet~\cite{DiffusionNet}, proposed by Sharp et al., is a neural network that learns on 3D surfaces by mimicking the intrinsic heat diffusion process. It diffuses per-vertex features across the surface based on the Laplace--Beltrami operator, which captures the intrinsic geometry of the manifold. The resulting heat operator acts as a geometry-aware smoothing filter that blends nearby features over time. In discrete form, DiffusionNet approximates this process using the cotangent Laplacian $L$ and mass matrix $M$, defined as
\begin{equation}
    h_t(u_0) = (M + tL)^{-1} M u_0,
\end{equation}
where $h_t(u_0)$ represents the diffused features after time $t$, followed by a lightweight MLPs for non-linearity. Because diffusion depends only on surface intrinsic geometry, the same learned weights transfer across meshes with different resolutions or triangulation, making the model compact, discretization-agnostic, and effective for tasks such as classification and regression on geometric data.

\subsection{Linear FACS Blendshape Rig}
\label{subsec:linearRig}
The linear FACS blendshape rig~\cite{Lewis2014pratice} models an animatable 3D face using a neutral mesh $M_0 = (V_0, F)$, where $V_0$ represents the vertex positions and $F$ the mesh connectivity. It also defines a set of $N$ blendshapes $\{M_i = (V_i, F)\}_{i=1}^{N}$, each obtained by adding a vertex offset $d_i$ to the neutral mesh: $V_i = V_0 + d_i$. Each blendshape corresponds to an Action Unit (AU) from the Facial Action Coding System (FACS)~\cite{ekman1978facial}, representing specific muscle movements such as ``Right Eye Close.'' Complex facial expression animation, involving the activation of multiple action units, is achieved by assigning a weight $w_i \in [0,1]$ to each blendshape and computing the final mesh $M = (V, F)$, where $ V = V_0 + \sum_{i=1}^{N} w_i d_i.$

\section{Method} 
\subsection{Data Collection}
\label{subsec:DataPreparation}
\begin{figure}
    \centering
    \includegraphics[width=0.99\linewidth]{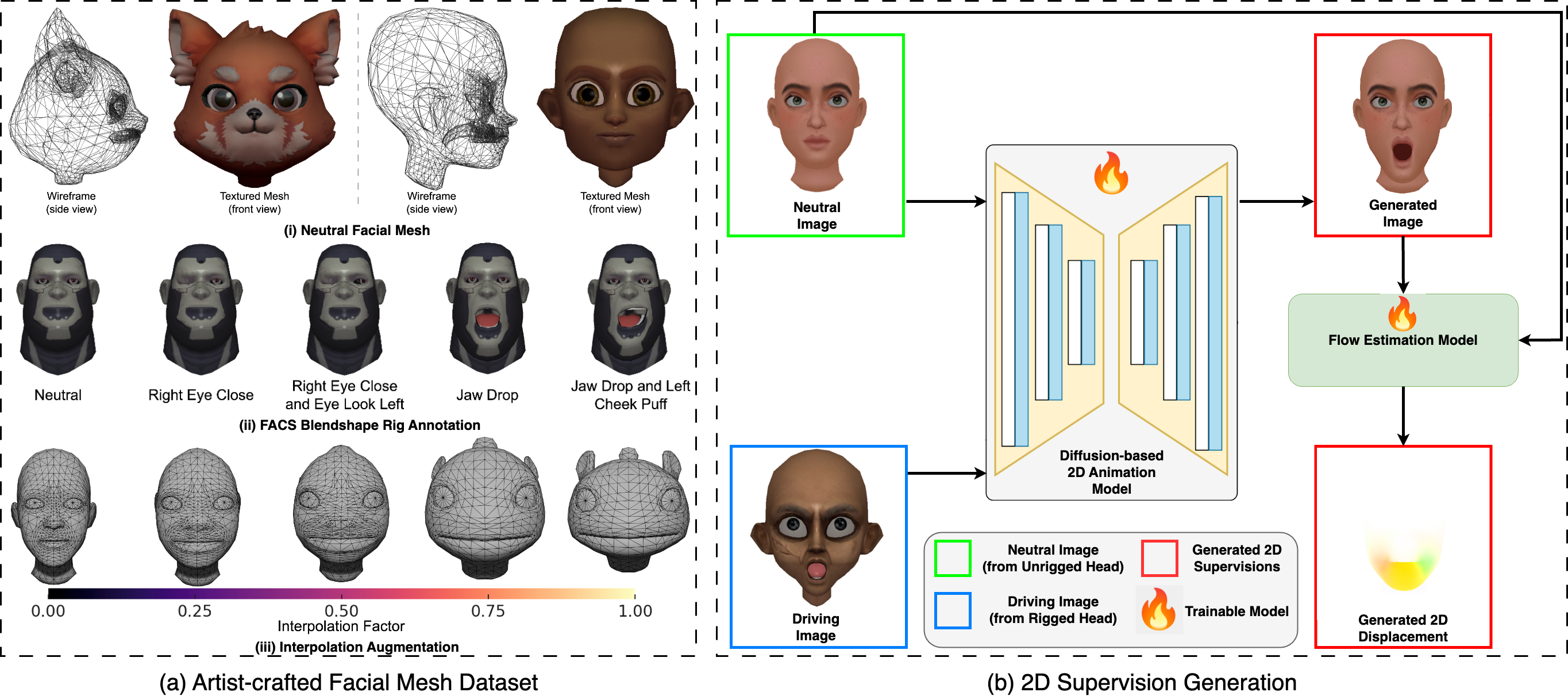}
    \caption{\textbf{(a)} Illustration of our artist-crafted facial mesh dataset. (i) Neutral head meshes from our dataset, each consisting of multiple disconnected components. (ii) A subset of neutral head meshes is meticulously annotated with blendshape rigs by professional artists. (iii) To augment the dataset, we develop a head interpolation strategy based on standardized UV layouts. \textbf{(b)} 2D Supervision Generation Pipeline: Given a posed image rendered from a rigged head and a neutral image from an unrigged head, the 2D animation model generates an image that replicates the expression in the posed image while preserving the identity of the neutral image. A flow estimation model is then applied to the neutral and generated posed images to predict the pixel offsets as 2D displacement.}
    \label{fig:datacollection}
    \vspace{-2mm}
\end{figure}
We collect a diverse set of artist-crafted facial meshes for model training and evaluation. As shown in Fig.~\ref{fig:datacollection} (a)(i), our dataset includes facial meshes with multiple disconnected components, such as separate eyeballs and features a variety of shapes, including both humanoid and non-humanoid heads.

Each dataset sample contains a neutral base mesh \(M_0\). For a select subset, artists manually annotated each mesh with its own complete blendshape rig \(\{M_i = (V_i, F)\}_{i=1}^{N}\) across \(N\) FACS training poses, as described in Sec.~\ref{subsec:linearRig} and illustrated in Fig.~\ref{fig:datacollection} (a)(ii). We set \(N = 96\), comprising 48 FACS poses and 48 corrective poses; further details are provided in the appendix. We also pair each blendshape with a one-hot-like FACS vector $A_i$ as pose representation, where activated action entries are set to 1. Furthermore, those heads were also annotated with facial landmarks specified as vertex indices. For unlabeled heads, only a neutral head mesh $M_0 = (V_0, F)$ is included.

Creating head meshes with complex rigs for animation is an expensive process.
In order to expand our dataset sufficiently for training a deep neural network, we developed a data augmentation strategy based on a standardized UV layout, enabling interpolate between different head geometries through linear blending to increases the size of our dataset, as illustrated in Fig.~\ref{fig:datacollection} (a)(iii).

\subsection{Deformation network}
\subsubsection{Network Architecture}
\begin{figure}
\centering
\includegraphics[width=0.9\linewidth]{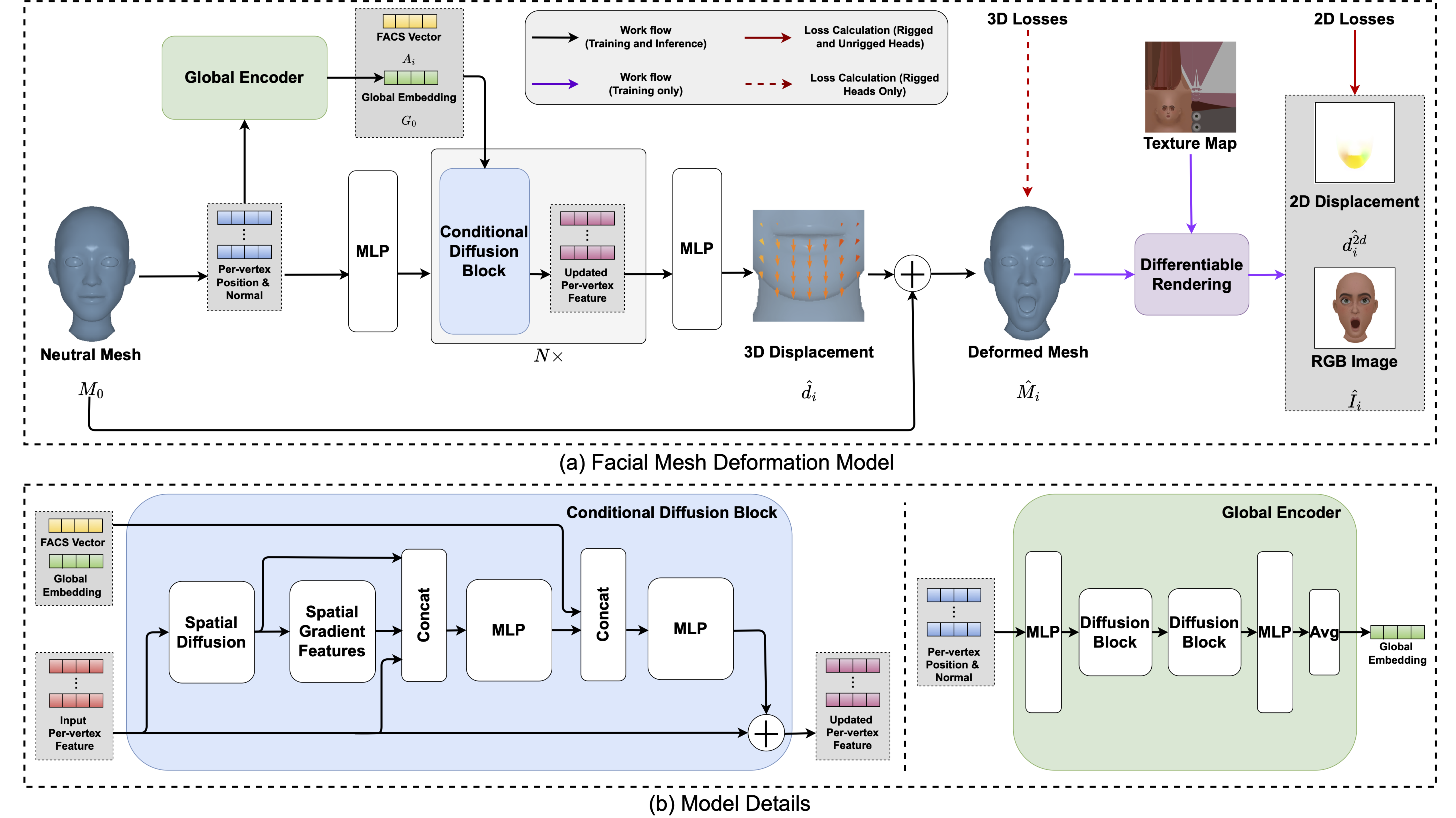}
\caption{Model Architecture. \textbf{(a)} Given a neutral facial mesh, our deformation model predicts the 3D displacement needed to deform the mesh into different expressions based on the input FACS vector. During training, 2D supervision is utilized for both rigged and unrigged heads, while 3D supervision is exclusively applied to rigged heads. \textbf{(b)} We modify the original diffusion block in DiffusionNet to support the FACS vector as an additional conditional inputs (left). Additionally, we design a global encoder that processes vertex positions and normals of the neutral facial mesh to capture holistic information across disconnected components (right).}
\label{fig:model}
\vspace{-4mm}
\end{figure}

As shown in Fig.~\ref{fig:model} (a), our deformation network takes the neutral facial mesh $M_0 = (V_0, F)$ and a FACS pose vector $A_i$ as inputs and predicts the displacement $\hat{d_i}$ required to deform the neutral mesh into the corresponding posed mesh $\hat{M_i} = (\hat{V_i}, F)$, where $\hat{V_i} = V_0 + \hat{d_i}$. The posed meshes obtained for all FACS poses together form a linear FACS blendshape rig.

We build our deformation network upon DiffusionNet~\cite{DiffusionNet} to take advantage of its triangulation-agonistic property. However, DiffusionNet struggles to handle multiple disconnected components as its diffusion mechanism cannot propagate information between them. Furthermore, it is limited to processing a single mesh without additional input. In our task, we aim to deform facial meshes with multiple disconnected components conditioned on an additional input: the FACS vector. To this end, we introduce two key modifications to the original DiffusionNet: (i) \textbf{Global Encoder} to capture holistic mesh characteristics across multiple disconnected components. As shown in the right of Fig.~\ref{fig:model} (b), this branch consists of a smaller 2-layer DiffusionNet that processes the input neutral mesh. Global average pooling is applied to the final layer's per-vertex features, producing a single vector encoding $G_0$ that compresses information about the mesh into a global feature vector. (ii) \textbf{FACS Conditioning:}  We modify the original diffusion block in DiffusionNet to integrate a FACS pose vector as a conditional input, guiding the network's generation of facial expressions. This allows the network to learn the relationship between FACS values and corresponding mesh deformations. As shown on the left of Fig.~\ref{fig:model} (b), the FACS pose vector $A_i$ is concatenated with the global feature vector $G_0$ to create a latent representation. This latent representation is then injected into each conditional diffusion block of the main network. Within each block, the latent vector is replicated across the vertex dimension and fused with the block's output features. This fused information is then processed by a small MLP to refine the mesh's latent features. 

\subsubsection{2D Supervisions for 3D Deformation Model}
\label{subsubsec:2DSupervision}
\begin{figure}[tb!]
    \centering
    \includegraphics[width=0.6\linewidth]{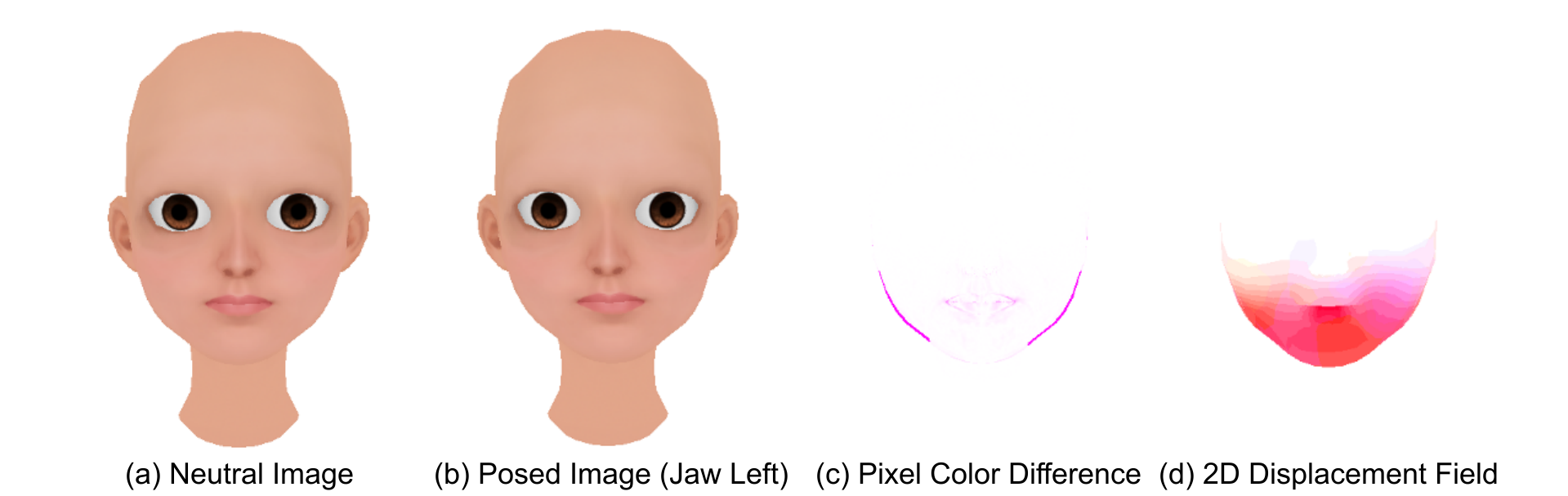}
    \caption{Illustration of our 2D displacement supervision (d), which provides denser feedback for the subtle pose differences between (a) and (b) than the appearance-level supervision (c). Subfigure (c) visualizes per-pixel color-difference magnitudes between (a) and (b), whereas subfigure (d) shows the corresponding pixel offsets using the standard optical-flow color map.}
    \label{fig:dis-2d}
    \vspace{-6mm}
\end{figure}
Relying solely on fully rigged heads limits the training dataset size due to the scarcity of high-quality 3D ground truth, which hampers generalization to unseen facial meshes. In contrast, 2D supervision is more readily available thanks to advancements in 2D generation models, enabling the inclusion of unrigged heads to scale up the training dataset to enhance generalization. Thus, we introduce 2D supervision for the face auto-rigging network in terms of appearance and motion variation. Specifically, for appearance, we use the front-view image and binary segmentation mask of the posed head as supervision. We render the RGB image $\hat{I_i}$ and binary mask $\hat{B_i}$ of the predicted mesh $\hat{M_i}$ onto the 2D image plane using differentiable rendering~\cite{SoftRasterizer,Pytorch3D}. The image loss $\mathcal{L}_{img}$ and mask loss $\mathcal{L}_{mask}$ are defined as the $l1$ distances between $\hat{I_i}$ with the ground-truth image $I_i$ and between $\hat{B_i}$ with the ground truth mask $B_i$, respectively.

Using appearance-level supervisions like image and mask losses, provides a straightforward way to optimize the 3D deformation network using 2D supervision. These losses offer strong supervisory signals for poses that result in significant changes in pixel's color value. However, many target FACS poses involve subtle expressions, where changes are less visually apparent. For instance, as shown in Fig.~\ref{fig:dis-2d}, comparing the neutral image in Fig.~\ref{fig:dis-2d} (a) with the jaw-left pose image in Fig.~\ref{fig:dis-2d} (b), the differences are barely noticeable to the human eye. Similarly, as illustrated in Fig.~\ref{fig:dis-2d} (c), the pixel error map on RGB value between these two images highlights that only a small portion of pixels contribute meaningful supervisory feedback for these subtle deformations. In other words, the magnitude of the loss remains minimal—even if the deformation model leaves all vertices fixed in the neutral expression.

To address this challenge, we introduce another 2D supervision for the 3D deformation model based on pixel motions. Specifically, we define the 2D displacement $d^{2d}_i$ as the offset of each pixel on the image plane between the neutral and posed images, analogous to optical flow. This 2D displacement is computed from the 3D displacement $d_i$ in a fully differentiable manner with differentiable rendering~\cite{SoftRasterizer, Pytorch3D} (see appendix for implementation). As shown in Fig.~\ref{fig:dis-2d} (d), the 2D displacement is more distinguishable for subtle facial expressions because it explicitly represents the motion of each pixel in 2D, rather than relying on RGB value changes. This is particularly beneficial in areas with uniform texture, such as cheek, where RGB value changes may be unnoticeable. We define the 2D displacement loss $\mathcal{L}_{dis-2d}$ as the $l2$ distance between the ground truth 2D displacement $d^{2d}_i$ and predicted 2D displacement $\hat{d^{2d}_i}$.

\subsubsection{2D Supervision Generation}
\label{subsubsec:2DGeneration}
For all rigged heads, we can obtain the above 2D supervisions by rendering from 3D ground truth. However, for unrigged heads, this is not feasible due to the absence of complete 3D ground truth deformations. To this end, we leverage recent advancements in 2D generation models to generate 2D supervision for unrigged heads. These models effectively distill appearance and motion priors from large-scale 2D image and video datasets, generalizing well across diverse scenarios. 

We implement a 2D face animation diffusion model based on Megactor~\cite{shurong2024megactor}. As illustrated in Fig.\ref{fig:datacollection} (b), this model takes a neutral reference image rendered from an unrigged head and a driving posed image rendered from a rigged head, animating the neutral image to replicate the expression in the posed image while preserving its identity. The generated images serve as image-based ground truth for unrigged heads during the training of the 3D deformation model. In practice, we select one rigged head, render all its FACS poses images, and use them as driving images to generate corresponding posed images for all unrigged heads. Ground truth masks are obtained using a traditional image segmentation model\cite{findcontour}, as all generated images are with a clean white background. 

For the 2D displacement, we use the optical flow estimation model RAFT~\cite{RAFT} to predict pixel offsets between the neutral image and the generated posed image of unrigged heads. These offsets serve as the ground truth 2D displacement for training the 3D deformation model.

To enhance the performance of the 2D face animation and flow estimation models on stylized faces in our artist-crafted dataset, we fine-tune their pre-trained weights using the ground truth renderings from a small set of rigged heads, improving effectiveness.

\subsubsection{Network Training and Inference}
\label{subsubsec:modeltraining}
We train the network in a two-stage, coarse-to-fine manner. In the first stage, the 3D deformation network is trained on a large-scale dataset comprising both rigged and unrigged heads, using only 2D supervision. We use a combination of photometric loss and 2D displacement loss, along with a $l_2$ regularization loss, $\mathcal{L}_{reg}$ on the predicted 3D displacement. This regularization loss helps to improve model convergence speed and prevent ``flying points" for non-line-of-sight vertices. The total training loss for the first stage is defined as:
\begin{equation} 
\mathcal{L}_{s1} = \alpha_1 \mathcal{L}_{img} + \alpha_2 \mathcal{L}_{mask} + \alpha_3 \mathcal{L}_{dis-2d},  + \alpha_4 \mathcal{L}_{reg} 
\end{equation}
where $\alpha$ are weighting parameters for different loss terms. 

In the second stage, we fine-tune the pretrained model from the first stage using only rigged heads, incorporating both 2D and 3D supervision to achieve high-precision deformation predictions. Since the 3D ground truth deformed mesh 
$M_i = (V_i, F)$ for FACS pose $i$ is available for rigged heads, we incorporate 3D supervision by applying the MSE loss $\mathcal{L}_{mse-3d}$ in 3D space between the ground truth and predicted mesh vertices $V_i$ and  $\hat{V_i}$.

For 2D supervision, in addition to the image loss and mask loss, we added two loss terms, landmark loss $\mathcal{L}_{lmk}$ and eye close loss $\mathcal{L}_{ec}$, as in~\cite{DECA} to provide supervision for specific facial landmarks and poses. We omit the 2D displacement loss in this stage since the 3D displacement ground truth is available. The total training loss for the second stage is defined as:
\begin{equation} 
\mathcal{L}_{s2} = \alpha_1 \mathcal{L}_{img} + \alpha_2 \mathcal{L}_{mask} + \alpha_3 \mathcal{L}_{mse-3d} + \alpha_4 \mathcal{L}_{lmk} + \alpha_5 \mathcal{L}_{ec}.
\end{equation}

The proposed model only consists of 5.4M parameters. Training runs on an instance with 8 NVIDIA A100 GPUs and takes about 2 days. For inference, it takes on average 8.72s on an Apple M2 Max CPU and 3.1s on an Nvidia T4 GPU to generate a FACS blendshape rig on the test set.

\section{Experiments}
In this section, we evaluate RAF on both the artist-crafted and in-the-wild facial meshes and compare it with the prior art NFR~\cite {NFR} and a representative deformation-transfer method~\cite {sumner2004deformation}.

\subsection{Evaluation on Artist-crafted Data }

\begin{table}[!hbt]
\vspace{-4mm}
  \centering
  \begin{minipage}[t]{0.48\linewidth}
    \centering
    \captionof{table}{Quantitative results on our artist-crafted dataset, validating each component of the model.}
    \resizebox{\linewidth}{!}{%
      \begin{tabular}{c |c|ccc|cc|cc}
        \toprule
         &\multicolumn{1}{c}{Network} &\multicolumn{3}{c}{Supervision terms} & \multicolumn{2}{c}{Training dataset} & \multicolumn{2}{c}{Test Results (mm)}  \\
         & Global encoder &$\mathcal{L}_{mse-3d}$ & $\mathcal{L}_{img}$ & $\mathcal{L}_{dis-2d}$ & Rigged & Unrigged & MAE $\downarrow$ & MAE Q95 $\downarrow$ \\\midrule
         w/o Global Encoder & \ding{55} & \ding{51} & \ding{55} & \ding{55}  &\ding{51} & \ding{55} &2.14 &6.64 \\
         w/o 2D Loss        & \ding{51} & \ding{51} & \ding{55} & \ding{55} &\ding{51} & \ding{55} &2.08 &5.84 \\
         w/o Unrigged Data  & \ding{51} & \ding{51} & \ding{51} & \ding{55} &\ding{51} & \ding{55} &2.01 &5.81 \\
         w/o 2D Displacement& \ding{51} & \ding{51} & \ding{51} & \ding{55} &\ding{51} & \ding{51} &1.95 &5.89 \\
         Full Model         & \ding{51} & \ding{51} & \ding{51} & \ding{51} &\ding{51} & \ding{51} &\textbf{1.92} &\textbf{5.63}\\
        \bottomrule
      \end{tabular}
    }%
    \label{tab:quantitative-internal}
  \end{minipage}
  \hfill
  \begin{minipage}[t]{0.48\linewidth}
    \centering
    \captionof{table}{Quantitative comparison with NFR and Deformation Transfer on 12 artist-annotated humanoid heads. (* additional inputs needed)}
    \resizebox{\linewidth}{!}{%
      \begin{tabular}{c|c|c}
        \toprule
         & MAE (mm) $\downarrow$ & MAE Q95 (mm) $\downarrow$ \\
         \midrule
        Deformation Transfer~\cite{sumner2004deformation}* & 2.93 & 8.41 \\
        \midrule
        NFR~\cite{NFR} & 2.77 & 7.21 \\
        Ours & \textbf{1.01} & \textbf{2.94} \\
        \bottomrule
      \end{tabular}
    }%
    \label{tab:val_compare}
  \end{minipage}
  \vspace{-3mm}
\end{table}

\begin{figure}[htb!]
\vspace{-3mm}
  \centering
  \includegraphics[width=0.88\linewidth]{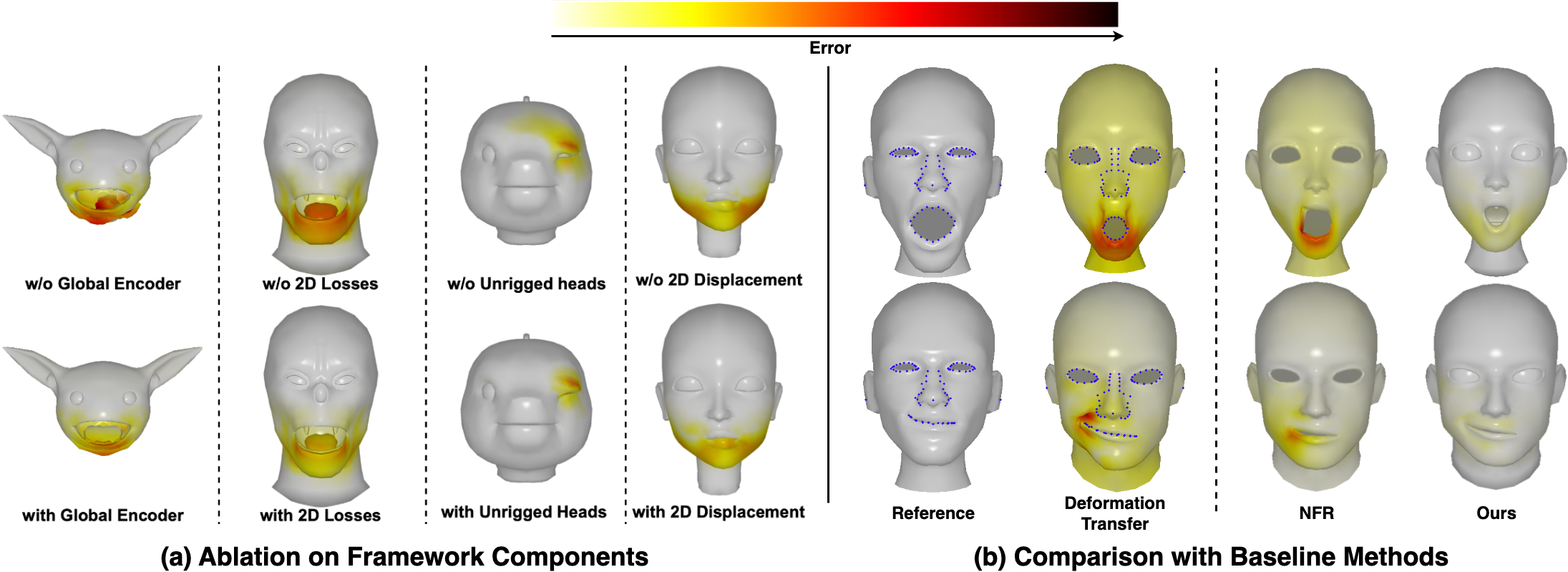}
  \caption{Visual comparisons. Meshes are colored by per‑vertex absolute error. \textbf{(a) Ablation on Framework Components.} 1st Col.: without the global encoder, disconnected parts intersect; 2nd Col.: 2D image loss reduces errors; 3rd Col.: additional unrigged heads improve generalization, addressing challenging cases such as animal eye closure; 4th Col.: 2D-displacement loss further refines subtle poses such as “Jaw Left.” \textbf{(b) Comparison with Baseline Methods.} Our method achieves more accurate and expressive animation results while handling multiple disconnected components. Reference mesh and corresponding points (marked as \textcolor{blue}{blue}) are provided for Deformation Transfer.}
  \label{fig:val_combined}
  \vspace{-2mm}
\end{figure}

We evaluate our model both quantitatively and qualitatively on our artist-crafted dataset. The evaluation includes two test sets: one with rigged heads for detailed accuracy analysis, and another with unrigged heads featuring diverse species and shapes to assess generalization on out-of-distribution samples, simulating real-world applications. 

For rigged heads with 3D ground-truth deformations, we compute the Mean Absolute vertex Error (MAE) and the 95th-percentile vertex error (MAE Q95) to capture challenging cases; both metrics are evaluated over the full set of 96 FACS poses. During evaluation, all facial meshes are normalized to fit within a unit sphere with a radius of 1 meter. Quantitative results are presented in Tab.~\ref{tab:quantitative-internal}, while qualitative results are shown in Fig.~\ref{fig:val_combined}(a). Together, these results validate the effectiveness of each component in our model. 

\begin{figure}[hbp!]
    \vspace{-2mm}
    \centering
    \resizebox{0.9\textwidth}{!}{
    \begin{minipage}[c]{0.5\textwidth}
        \centering
        \captionof{table}{Ablation on the global encoder.}
        \resizebox{\textwidth}{!}{%
        \begin{tabular}{ccccc}
            \toprule
            \textbf{Global Encoder} & \textbf{All Other Components} & \textbf{MAE $\downarrow$} & \textbf{MAE Q95 $\downarrow$} & \textbf{Penetration $\downarrow$} \\
            \midrule
            $\times$ & $\times$ & 2.14 & 6.64 & 0.377 \\
            $\times$ & \checkmark & 2.16 & 6.08 & 0.405 \\
            \checkmark & $\times$ & 2.08 & 5.84 & \textbf{0.166} \\
            \checkmark & \checkmark & \textbf{1.92} & \textbf{5.63} & 0.173 \\
            \bottomrule
        \end{tabular}%
        }
        \label{tab:ablation_global}
    \end{minipage}
    \hfill
    \begin{minipage}[c]{0.45\linewidth}
        \centering
        \includegraphics[width=\textwidth]{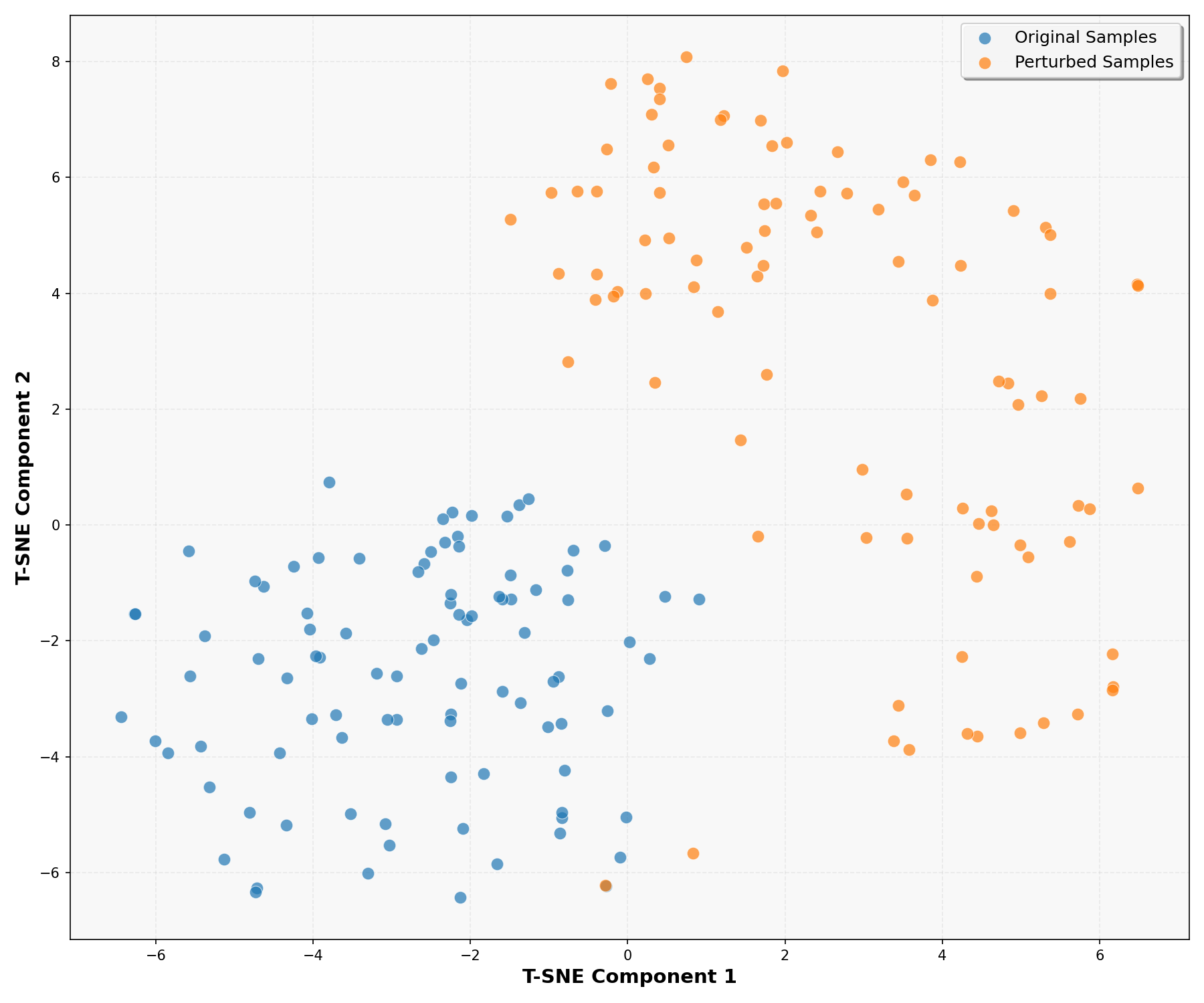}
        \caption{t-SNE visualization of features from the global encoder.}
        \label{fig:tsne}
    \end{minipage}
    }
    \vspace{-2mm}
\end{figure}

We also conduct additional ablations on the global encoder to demonstrate how a single feature vector from it enables our model to handle multiple disconnected components. We evaluate penetration between inner components (e.g., teeth) and the outer face surface by reporting the percentage of penetrating vertices with and without the global feature (Tab.~\ref{tab:ablation_global}). Furthermore, we perturb the disconnected components by randomly adding offsets or removing varying numbers of components from each sample. As shown in Fig.~\ref{fig:tsne}, the t-SNE visualization of global features from the perturbed samples forms separate clusters from the original ones. These results demonstrate that the global feature effectively encodes both the position and presence of disconnected components, avoiding penetration and achieving accurate deformation.

To ensure a fair comparison with NFR~\cite{NFR}, each input mesh was preprocessed following the same procedure as in their original implementation, to retain only the largest connected component of the neutral mesh with auxiliary structures (e.g., eyeballs and the mouth socket) removed and the inner surfaces of the lips and eyelids trimmed. 
Since NFR is trained and evaluated only on humanoid faces, we limit the test cases to 12 humanoid heads. Another method that we compare to is the Deformation Transfer~\cite{sumner2004deformation}, which requires an exemplar expression mesh and user-annotated point correspondences as additional input. 
We choose one rigged head from the training set as the exemplar, deform it into all FACS poses, and provide
artist-annotated landmarks as correspondence points. As reported in Tab. \ref{tab:val_compare} and Fig.~\ref{fig:val_combined}(b), our method outperforms both baselines by a wide margin. It also has the additional advantage of not requiring any additional input and being able to handle meshes with multiple disconnected components.

\begin{figure}[hb!]
    \vspace{-1mm}
    \centering
    \includegraphics[width=0.7\linewidth]{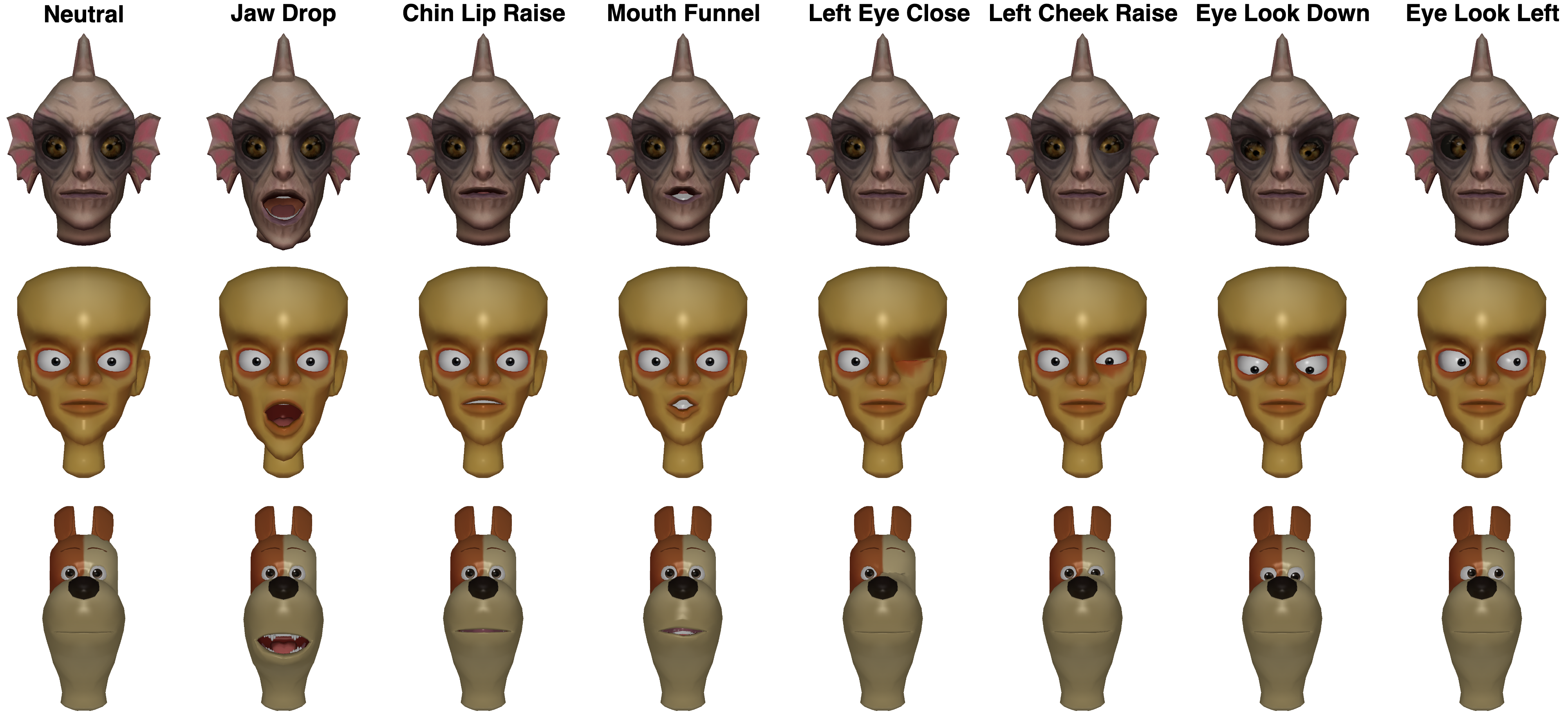}
    \caption{Qualitative results on our artist-crafted unrigged heads.}
    \label{fig:internal_heads}
    \vspace{-3mm}
\end{figure}

For the unrigged head test set, we provide qualitative results only, as 3D ground truth data is unavailable. Fig.~\ref{fig:internal_heads} showcases qualitative examples, where our model delivers highly accurate and vivid auto-rigging results across facial meshes of various shapes and styles. Additional results for all FACS poses and more samples can be found in the supplementary materials.  

\begin{figure}
\vspace{-3mm}
    \centering
    \includegraphics[width=0.7\linewidth]{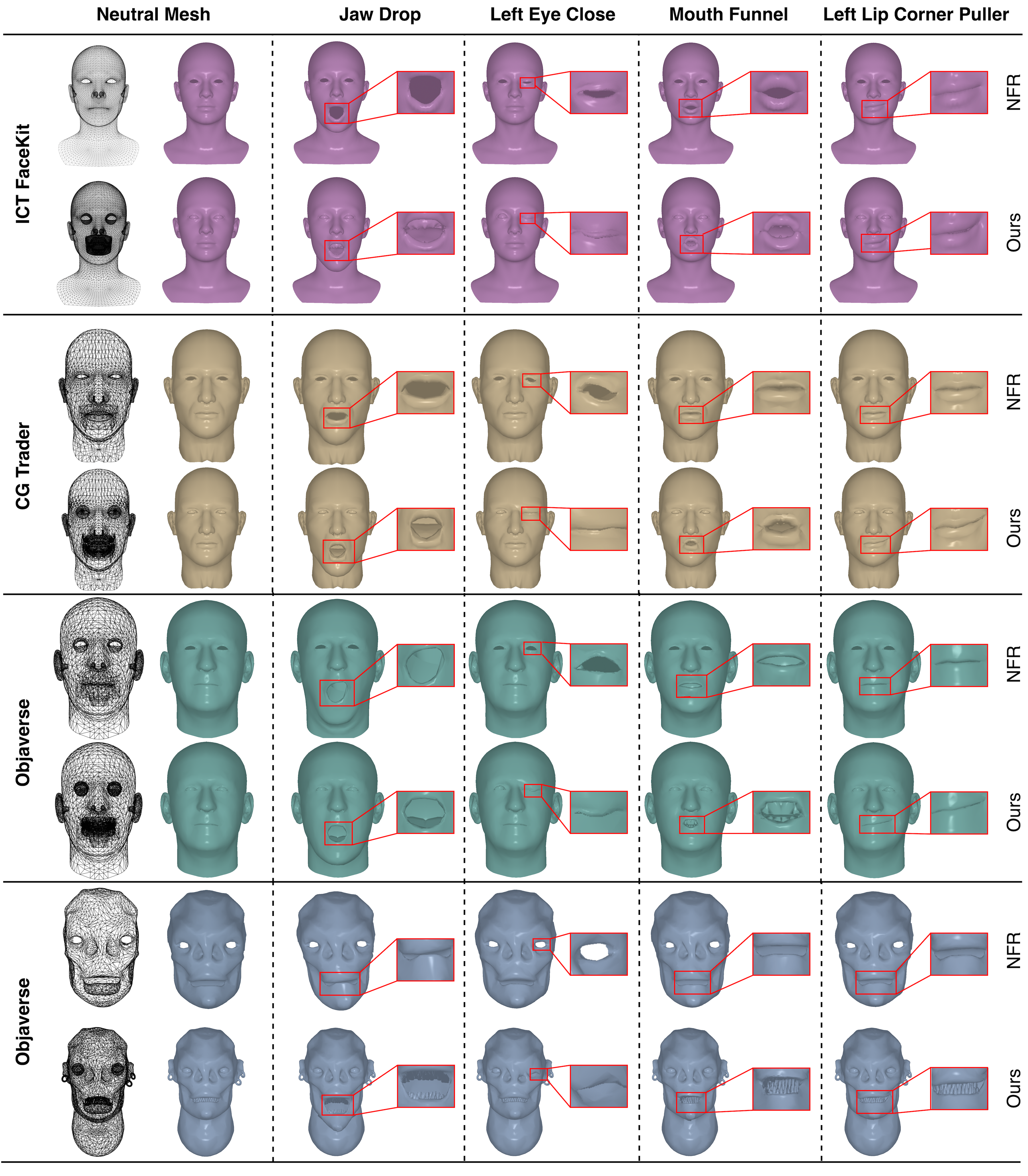}
    \caption{Auto-rigging results on in-the-wild facial meshes compared with NFR~\cite{NFR}.}
     \label{fig:wild-head}
     \vspace{-6mm}
\end{figure}

\subsection{Evaluation on In-the-wild Heads}
Our method generalizes effectively to in-the-wild facial meshes with diverse topology and shape variations. To demonstrate this, we present qualitative results on samples from ICT FaceKit~\cite{ictface}, Objaverse~\cite{Objectverse, Objectverse-xl}, and CGTrader~\cite{cgtrader2025}, and compare that with the results by NFR~\cite{NFR}. We do not provide a comparison with Deformation Transfer here, as correspondence point annotations are not available for these samples.
We similarly preprocess the input meshes for the comparison with NFR. As shown in Fig.~\ref{fig:wild-head}, our model consistently achieves better accuracy and generalizability. In particular, although NFR was trained on the ICTFaceKit dataset and ours was not, our results are comparable to those of NFR.
For humanoid assets from Objaverse and CGTrader, neither our method nor NFR was trained on data from these sources, our model demonstrates superior performance. For the non-humanoid head (last column), NFR leaves it largely undeformed, whereas our model successfully generalizes to this challenging case.

\begin{figure}
    \centering
    \includegraphics[width=0.7\linewidth]{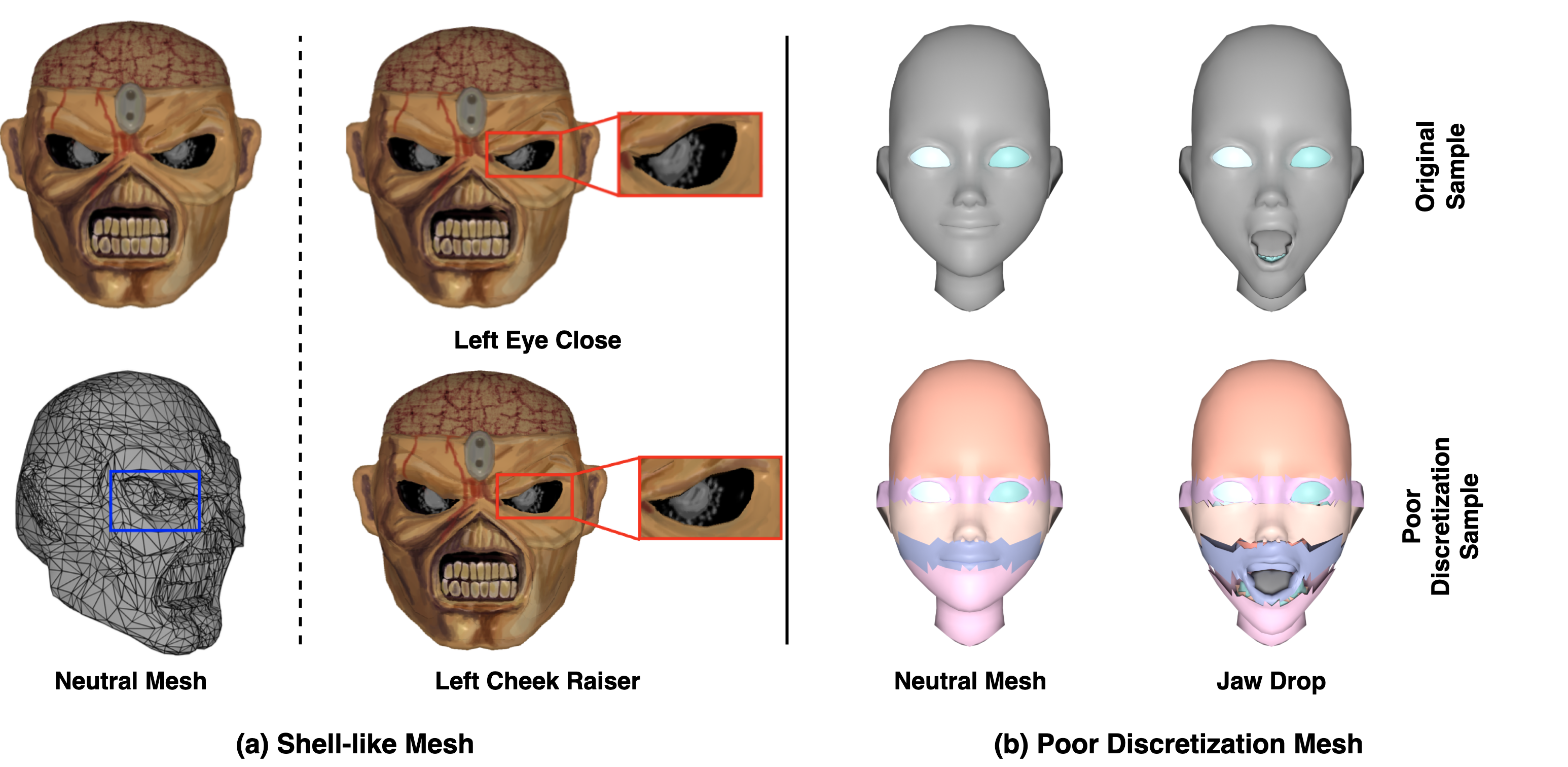}
    \caption{Failure cases of our RAF model.}
     \label{fig:fail}
     \vspace{-2mm}
\end{figure}

\subsection{Applications}
We demonstrate three real-world applications of RAF: (i) user-controlled animation, where the predicted FACS rig allows users to pose a mesh by editing FACS parameters; (ii) video-to-mesh retargeting, which transfers expressions of a subject in the video via tracked FACS sequences to an unrigged mesh; (iii) 
animating a facial mesh generated from a text-to-3D model, turning it from a neutral facial mesh into a fully animatable avatar. Demos can be found on our project page.

\section{Conclusion}
We propose RAF, a framework for auto-rigging facial meshes. Powered by our tailored design for multiple disconnected components and FACS conditioning and scaled by training on unrigged heads with 2D supervision, RAF can animate meshes of diverse topologies with even multiple disconnected components, across both artist-crafted assets and in-the-wild samples.

\textit{Limitations and Future Work.} Our model faces challenges in two scenarios: (i) When the input mesh structure deviates significantly from the training data, such as shell-like meshes that lack the fine-grained geometric details necessary for high-quality facial animation, the model's performance may decline (Fig.\ref{fig:fail} (a)). Expanding the dataset to include a broader range of mesh structures could enhance generalization in such cases. (ii) When the mesh has poor discretization that causes the main facial mesh to break into multiple disconnected components (shown in different colors), our model fails to maintain spatial coherence among these components after deformation (Fig.\ref{fig:fail} (b)). Incorporating a diffusion operator defined on a high-quality background triangulation~\cite{sharp2019navigating} could enhance robustness in such cases.

\section{Acknowledgement}
We thank Hsueh-Ti Derek Liu, Chrystiano Araújo, and Jinseok Bae for proofreading the draft and providing helpful comments, and Jihyun Yoon for curating the dataset. Wenchao Ma is supported by a travel grant from the Penn State College of Information Sciences and Technology for attending the conference.

\newpage
\small{
\bibliographystyle{plain}
\bibliography{main}
}

\newpage
\appendix
{\Large \textit{Appendix}}
\section{Training and Inference Details}
In the first stage of training, the weights for the image loss, mask loss, 2D displacement loss, and regularization loss are set to $10.0$, $1.0$, $1.0$, and $0.0001$, respectively. In the second stage, the weights for the image loss, mask loss, 3D MSE loss, 2D landmark loss, and 2D eye closure loss are set to $10.0$, $1.0$, $100.0$, $0.5$, and $0.5$, respectively. We train our model on an Nvidia A100 instance with 8 GPUs and a total batch size of 8 (i.e., effectively 1 sample per GPU if using distributed data parallel). The training proceeds in two stages. For the first stage, we train the deformation model on both rigged and unrigged head datasets (8,386 samples in total) using only 2D supervision for 15 epochs. This stage typically takes around 1.5 days to complete. For the second stage, we then finetune the model from the first stage on the rigged head dataset (2,929 samples), incorporating both 2D and 3D supervision for 20 epochs. This finetuning phase finishes in approximately 1 day. Throughout both stages, we use the Adam optimizer, initializing the learning rate at 0.0001. For learning rate scheduling, we employ CosineAnnealingWarmRestarts, allowing it to decay from 0.0001 to nearly 0 by the end of training. Additionally, we use a warm-up phase of 20,000 steps to stabilize early training. 

For inference speed, our model runs a single forward pass to predict blendshapes offline, requiring only one run per input mesh. The outputs are converted into classical FACS blendshape rigs, enabling efficient animation by simply linear blending. The proposed model consists of 5.4M parameters and it takes on average 8.72s on an Apple M2 Max CPU and 3.1s on a Nvidia T4 GPU to generate a FACS blendshape rig on the test set (1,750 vertices, 3,362 faces on average).

\section{Details for 2D Displacement Calculation}
In the following code sample, we demonstrate how to compute the 2D displacement of each pixel from mesh vertex deformations in a fully differentiable manner. This implementation leverages PyTorch3D’s differentiable rendering functionality.

\begin{lstlisting}
def render_displacement(vertices, deformed_vertices, faces, renderer, camera, res=(512,512)):
    """
    Parameters
    ----------
    vertices: torch.tensor (V, 3)
    deformed_vertices: torch.tensor (V, 3)
    faces: torch.tensor (F, 3)
    renderer: pytorch3d.renderer.MeshRenderer object
    camera: pytorch3d.renderer.cameras.CamerasBase object
    res: tuple

    Returns
    -------
    displacement_2D: torch.tensor (res[0], res[1], 2)
    """

    verts_2d = camera.transform_points_screen(vertices, image_size=res)
    verts_2d_deformed = camera.transform_points_screen(deformed_vertices, image_size=res)
    verts_flow = (verts_2d_deformed - verts_2d)[:, :2]  # Vx2
    verts_flow = verts_flow / res * 0.5 + 0.5  # 0~1
    flow_tex = torch.nn.functional.pad(verts_flow, pad=[0, 1])  # Vx3
    texture = TexturesVertex(verts_features=[flow_tex])
    meshes = pytorch3d.structures.Meshes(
        verts=[vertices], faces=[faces], textures=texture
    )
    displacement_2D = renderer(meshes, cameras=camera)

    return displacement_2D[...,:2].squeeze()
\end{lstlisting}

\section{Effectiveness of 2D Generation Pipeline}
\begin{figure}[htb!]
    \centering
    \includegraphics[width=\linewidth]{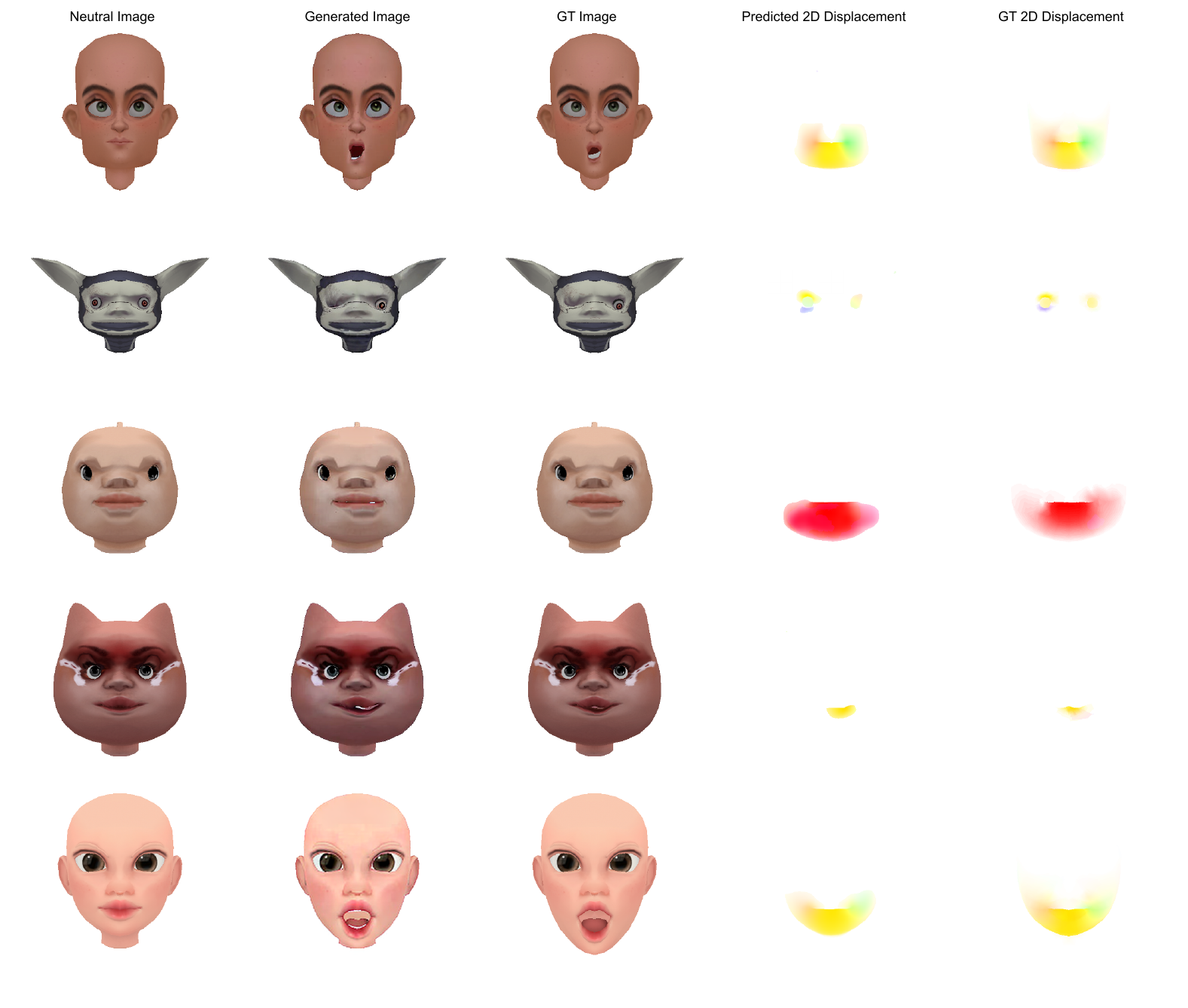}
    \caption{Example results of 2D generation pipeline.}
    \label{fig:2dgen_res}
\end{figure}

To validate the effectiveness of our 2D supervision generation pipeline, we exclude several rigged heads during the fine-tuning of the 2D face animation model and the flow estimation model. In Fig.~\ref{fig:2dgen_res}, we present random sample results showcasing different faces and poses. The ground truth images and 2D displacements are rendered using the ground truth deformations of the rigged heads. The 2D face animation model generates pose images based on the neutral image input, while the flow estimation model takes the neutral and generated images as input to predict the 2D displacement. The 2D displacement is visualized according to the standard optical flow convention.

\section{Data Collection Details}

\begin{figure*}
    \centering
    \includegraphics[width=\linewidth]{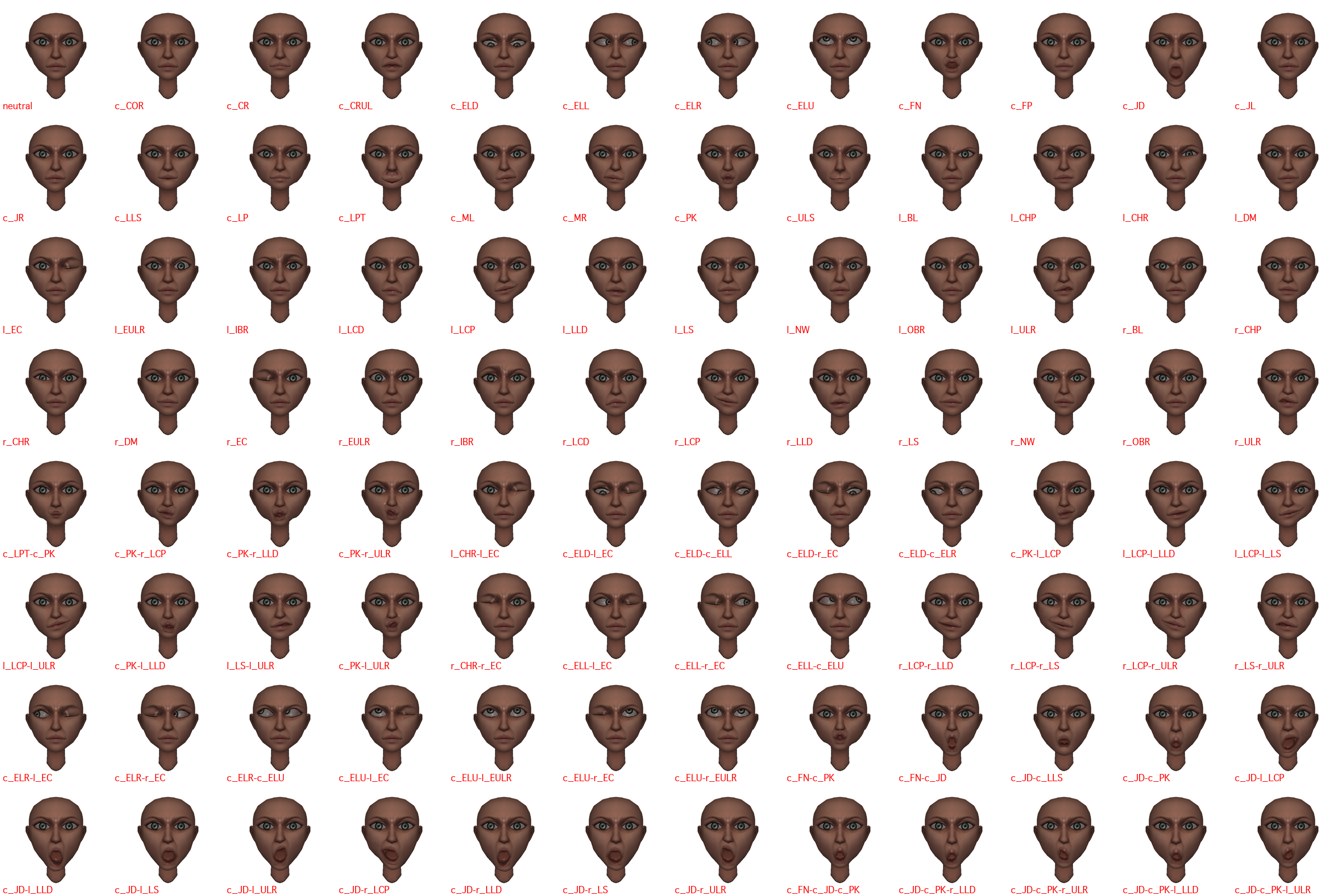}
    \caption{A sample of all the FACS and corrective poses used in this work.}
    \label{fig:facs}
\end{figure*}

\textbf{FACS Poses} For each rigged heads, our artist annotated 48 FACS poses and 48 corrective poses as blendshape rig. We show those 48 FACS poses in Tab.~\ref{tab:map}, and Fig~\ref{fig:facs}. In addition to blendshapes for individual FACS poses, we generate corrective blendshapes by linearly combining certain poses and manually correcting artifacts. These corrective blendshapes account for the complex deformations resulting from pose interactions. 

\textbf{Semantic Annotation} We provide a semantic annotation map for rigged heads, labeling different regions on the mesh (e.g., ears, mouth, eyes), along with facial landmark annotations specified as vertex indices. These annotations allow for the application of weighted losses or region-specific training objectives.

\textbf{Head Interpolation} First, we standardized the UV layout across all head meshes, ensuring that corresponding facial features like eyes and mouths occupy the same region in UV space. This consistent mapping enables the identification of 3D correspondences between vertices on different meshes. Using these correspondences, we can smoothly interpolate between different head geometries through linear blending to significantly increases the size of our dataset.

\begin{table}[!ht]
\centering
\renewcommand{\arraystretch}{1.2}
\resizebox{\linewidth}{!}{
\begin{tabular}{|c|l|l|c|l|l|}
\hline
\# & \textbf{SHORT} & \textbf{FULL} & \# & \textbf{SHORT} & \textbf{FULL} \\
\hline
1  & neutral & neutral                   & 25 & l\_EC   & LeftEyeClosed               \\
2  & c\_COR  & Corrugator                & 26 & l\_EULR & LeftEyeUpperLidRaiser       \\
3  & c\_CR   & ChinRaiser                & 27 & l\_IBR  & LeftInnerBrowRaiser         \\
4  & c\_CRUL & ChinRaiserUpperLip        & 28 & l\_LCD  & LeftLipCornerDown           \\
5  & c\_ELD  & EyesLookDown              & 29 & l\_LCP  & LeftLipCornerPuller         \\
6  & c\_ELL  & EyesLookLeft              & 30 & l\_LLD  & LeftLowerLipDepressor       \\
7  & c\_ELR  & EyesLookRight             & 31 & l\_LS   & LeftLipStretcher            \\
8  & c\_ELU  & EyesLookUp                & 32 & l\_NW   & LeftNoseWrinkler            \\
9  & c\_FN   & Funneler                  & 33 & l\_OBR  & LeftOuterBrowRaiser         \\
10 & c\_FP   & FlatPucker                & 34 & l\_ULR  & LeftUpperLipRaiser          \\
11 & c\_JD   & JawDrop                   & 35 & r\_BL   & RightBrowLowerer            \\
12 & c\_JL   & JawLeft                   & 36 & r\_CHP  & RightCheekPuff              \\
13 & c\_JR   & JawRight                  & 37 & r\_CHR  & RightCheekRaiser            \\
14 & c\_LLS  & LowerLipSuck              & 38 & r\_DM   & RightDimpler                \\
15 & c\_LP   & LipPresser                & 39 & r\_EC   & RightEyeClosed              \\
16 & c\_LPT  & LipsTogether              & 40 & r\_EULR & RightEyeUpperLidRaiser      \\
17 & c\_ML   & MouthLeft                 & 41 & r\_IBR  & RightInnerBrowRaiser        \\
18 & c\_MR   & MouthRight                & 42 & r\_LCD  & RightLipCornerDown          \\
19 & c\_PK   & Pucker                    & 43 & r\_LCP  & RightLipCornerPuller        \\
20 & c\_ULS  & UpperLipSuck              & 44 & r\_LLD  & RightLowerLipDepressor      \\
21 & l\_BL   & LeftBrowLowerer           & 45 & r\_LS   & RightLipStretcher           \\
22 & l\_CHP  & LeftCheekPuff             & 46 & r\_NW   & RightNoseWrinkler           \\
23 & l\_CHR  & LeftCheekRaiser           & 47 & r\_OBR  & RightOuterBrowRaiser        \\
24 & l\_DM   & LeftDimpler               & 48 & r\_ULR  & RightUpperLipRaiser         \\
\hline
\end{tabular}
}
\caption{FACS Short and Full Name Mapping.}
\label{tab:map}
\end{table}

\section{Dataset Split} \label{sec:dataset_split} 
Our dataset includes 161 rigged heads and 175 unrigged heads. From these, a subset of 24 rigged heads with 3D ground-truth annotations forms the test set to for accurate absolute error evaluation. Additionally, we select 37 diverse unrigged heads as the test set, representing different species and shapes to evaluate the model's generalization on out-of-distribution (OOD) faces.  For training, we augment the dataset using interpolations, manually filtering out poor interpolation results. Specifically, we interpolate the remaining 137 unrigged heads with a factor of 50, generating 5,457 samples, and interpolate the remaining 137 rigged heads with a factor of 25, producing 2,929 samples. 

\section{Pre-processing for Baseline Method NFR}
\begin{figure}
    \centering
    \includegraphics[width=0.6\linewidth]{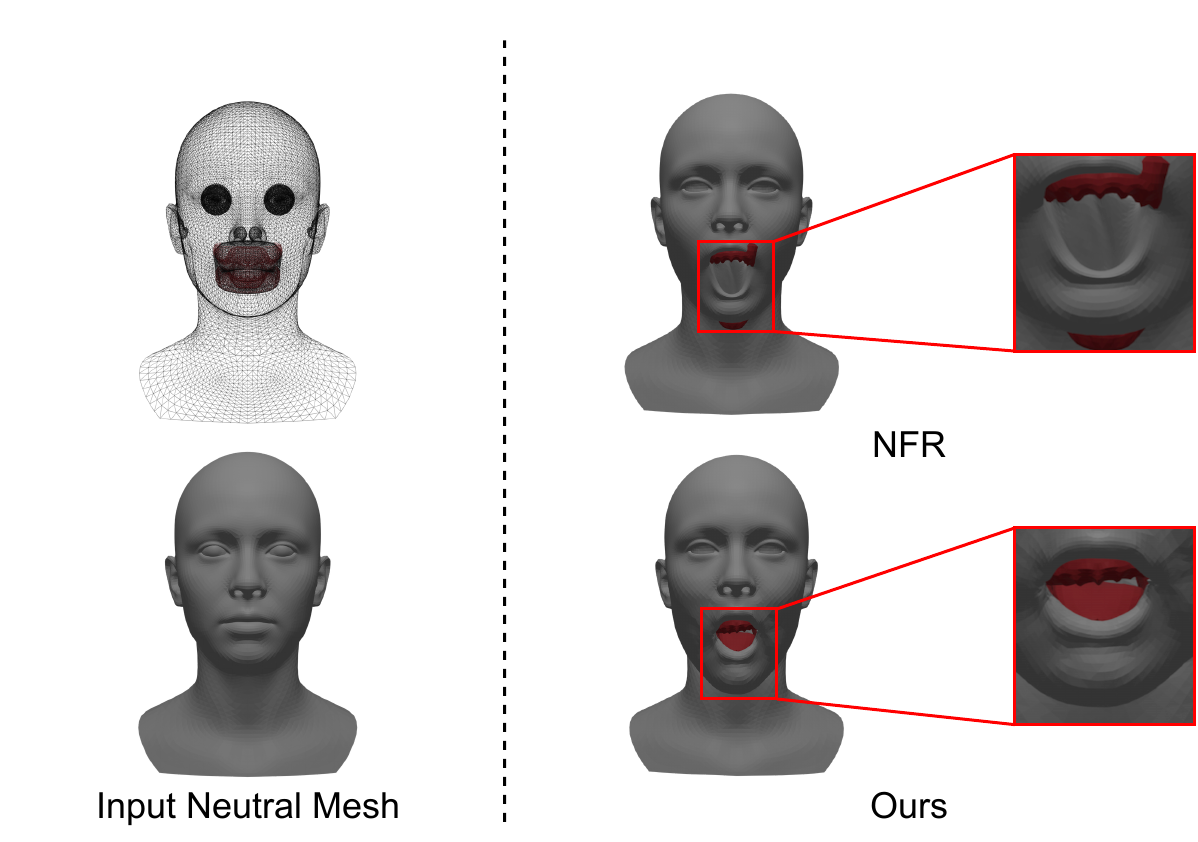}
    \caption{Compared to NFR during inference on meshes with multiple disconnected components from ICT Facekit Dataset. We highlight one of these components: "gums and tongue" in red. While animating a Jaw Drop pose, this component causes penetration issues for NFR.}
    \label{fig:multi-cc}
\end{figure}

\begin{figure}
    \centering
    \includegraphics[width=0.5\linewidth]{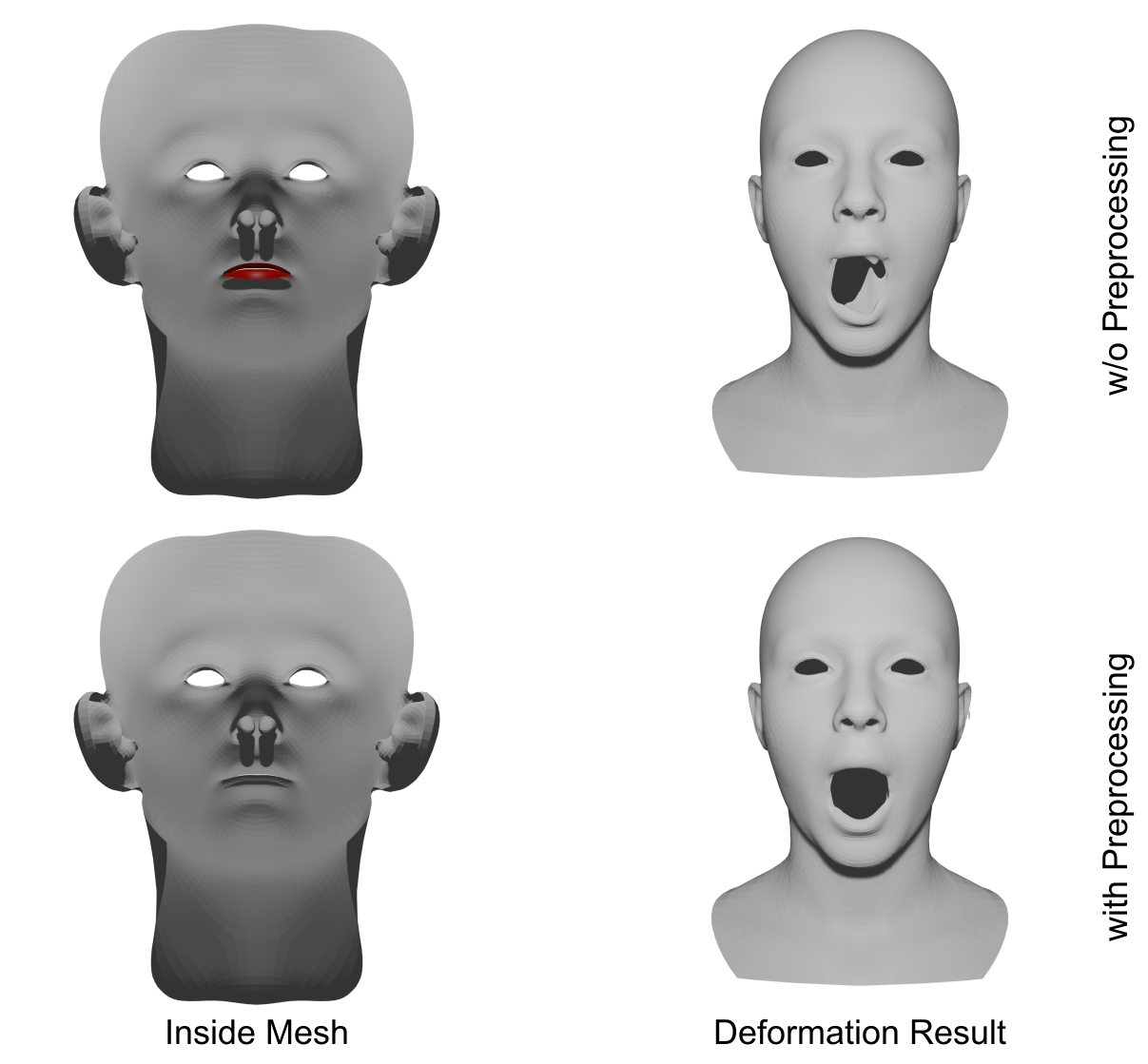}
    \caption{Illustration of the trimming preprocessing step for NFR. The inner‑lip surfaces to be trimmed are highlighted in \textcolor{red}{red} in the top‑left figure. Omitting this step results in implausible deformations produced by NFR.}
    \label{fig:trim}
\end{figure}

All NFR baseline results were obtained after applying the official preprocessing pipeline\footnote{\url{https://github.com/dafei-qin/NFR_pytorch}}: we keep only the largest connected component and remove the inner-lip and eyelid surfaces. These steps are crucial for NFR to generate reasonable deformations. Figure~\ref{fig:multi-cc} shows that retaining multiple disconnected components causes self-penetration, while Fig~\ref{fig:trim} shows jarring artifacts when the inner-lip surfaces are not trimmed. In contrast, our method do not need such preprocessing.

\section{Border Impact}
Our face-autorigging framework could broaden access to high-quality animation by letting small studios, educators, and assistive-tech developers create expressive avatars quickly, which benefits entertainment, remote communication, and certain medical visualization tasks. However, the same ease of use can lower the barrier for deepfake production, intensifying privacy concerns around emotion tracking and biometric profiling. Careful dataset curation, explicit usage licenses, and watermarking tools are essential to realize the creative upside while limiting misuse and inequitable impacts.

\end{document}